\documentclass[10pt,twocolumn,letterpaper]{article}
\usepackage{iccv}
\usepackage{times}
\usepackage{epsfig}
\usepackage{graphicx}
\usepackage{amsmath}
\usepackage{amssymb}

\usepackage{eso-pic}
\usepackage[latin9]{inputenc}
\usepackage{array}
\usepackage{multirow}
\PassOptionsToPackage{normalem}{ulem}
\usepackage{ulem}

\usepackage[breaklinks=true,bookmarks=false,colorlinks,linkcolor=red,anchorcolor=blue, citecolor=green]{hyperref}

\iccvfinalcopy 

\def\iccvPaperID{****} 

\ificcvfinal\fi

\begin{document}

\title{Sparse-to-dense Feature Matching: Intra and Inter domain Cross-modal Learning in Domain Adaptation for 3D Semantic Segmentation}

\author{Duo Peng$^{1}$\quad Yinjie Lei$^{1,}$\thanks{Corresponding Author: Yinjie Lei (yinjie@scu.edu.cn)}\quad Wen Li$^{2}$\quad Pingping Zhang$^{3}$\quad Yulan Guo$^{4}$\\
$^{1}$Sichuan University\quad $^{2}$University of Electronic Science and Technology of China\\
$^{3}$Dalian University of Technology\quad $^{4}$National University of Defense Technology\\
{\tt\small duo\_peng@stu.scu.edu.cn}, {\tt\small yinjie@scu.edu.cn}, {\tt\small liwenbnu@gmail.com}\\ 
{\tt\small zhpp@dlut.edu.cn}, {\tt\small yulan.guo@nudt.edu.cn}
}


\maketitle
\ificcvfinal\thispagestyle{empty}\fi

\begin{abstract}
  Domain adaptation is critical for success when confronting with the lack of annotations in a new domain. As the huge time consumption of labeling process on 3D point cloud, domain adaptation for 3D semantic segmentation is of great expectation. With the rise of multi-modal datasets, large amount of 2D images are accessible besides 3D point clouds. In light of this, we propose to further leverage 2D data for 3D domain adaptation by intra and inter domain cross modal learning. As for intra-domain cross modal learning, most existing works sample the dense 2D pixel-wise features into the same size with sparse 3D point-wise features, resulting in the abandon of numerous useful 2D features. To address this problem, we propose \textbf{D}ynamic \textbf{s}parse-to-dense \textbf{C}ross \textbf{M}odal \textbf{L}earning (DsCML) to increase the sufficiency of multi-modality information interaction for domain adaptation. For inter-domain cross modal learning, we further advance \textbf{C}ross \textbf{M}odal \textbf{A}dversarial \textbf{L}earning (CMAL) on 2D and 3D data which contains different semantic content aiming to promote high-level modal complementarity. We evaluate our model under various multi-modality domain adaptation settings including day-to-night, country-to-country and dataset-to-dataset, brings large improvements over both uni-modal and multi-modal domain adaptation methods on all settings. Code is available at \href{https://github.com/leolyj/DsCML}{https://github.com/leolyj/DsCML}
\end{abstract}



\section{Introduction}

    \begin{figure}[htp]
    \centering{}\vspace{-2mm}
     \includegraphics[scale=0.18]{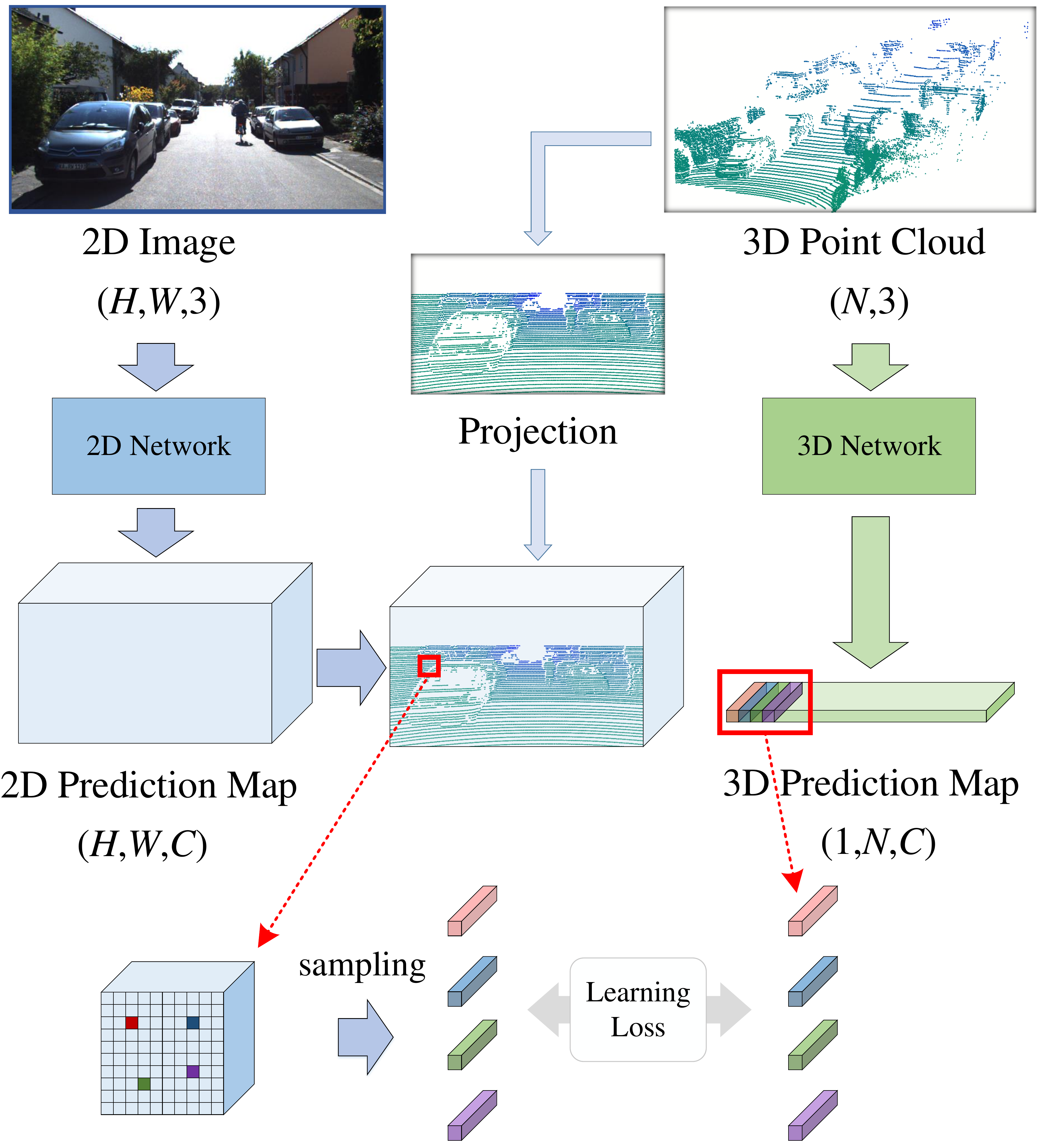} \caption{The common strategy of feature processing for 2D-3D cross modal learning. The 2D dense feature map with dense pixel-wise features are sampled to sparse features with the same size of 3D point features. As a result, such sparse-to-sparse feature matching only leverages quite limited 2D features and might cause insufficient 2D-3D information interaction. Specifically, $H$ and $W$ are the height and width of 2D image respectively. $N$ denotes the number of points in point cloud. $C$ is the number of categories for semantic segmentation.}
    \label{--intro}\vspace{-5mm}
    \end{figure}

    3D semantic segmentation is a challenging task with plenty of real-world applications, such as particular robotics, autonomous driving and virtual reality. Like other tasks of scene perception, 3D semantic segmentation also faces the challenge of domain shift. For instance, training on one country and testing on another and in different times of the day may lead to significantly dreadful performance. Plenty of domain adaptation methods are proposed to address such domain shift on the task of 2D semantic segmentation \cite{luo2019taking,hoffman2016fcns,hoffman2018cycada,peng2021global,lei2020hierarchical,yang2020label} but rarely on 3D \cite{wu2019squeezesegv2}.

    In recent works for multi-modality datasets creation, researchers often incidentally capture 2D images as counterpart when executing the data collection of 3D point clouds. In light of this, Jaritz \emph{et al}. \cite{jaritz2020xmuda} proposes a cross modal learning method to address domain adaptation for 3D semantic segmentation. Through the complementary advantages between 2D and 3D data, the multi-modality domain adaptation can bring large improvements over uni-modal adaptation methods. And the 2D images are leveraged without labels which means no additional human efforts in labeling work.

    While this work \cite{jaritz2020xmuda} already explored multi-modality in domain adaptation, as shown in Fig. \ref{--intro}, it executes the learning between 2D and 3D only towards the matched features by projection from 3D points to 2D image. Multitudes of mismatched features which also contain useful information are discarded. We consider whether 2D and 3D features can be more sufficiently utilized in cross modal learning. This is challenging as the two inputs are heterogeneous and contain different number of elements. Compared to the sparse 3D point clouds, pixels in 2D image are dense and with compact layout. Even if all features of 3D point cloud well matched with the corresponding 2D pixel-wise features, there can be still lots of 2D features are mismatched. That is why numerous works of other perception tasks \cite{ren2016faster,de2018robust,kumar2020lidar,chiang2019unified,jaritz2019multi,lei2019deep,su2018splatnet} also capitalize on multi-modality in the same way with \cite{jaritz2020xmuda}, i.e., sparse-to-sparse feature matching.

    To address such limitation, we propose a strategy namely \textbf{D}ynamic \textbf{s}parse-to-dense \textbf{C}ross \textbf{M}odal \textbf{L}earning (DsCML) where the sparse point cloud features and dense pixel features can sufficiently interact with each other. Specifically, the proposed DsCML is inspired from the fact that in 2D semantic segmentation, the neighboring pixels are mostly classified to a same category. And all the same-categorized pixel features should be sampled to exchange information with the corresponding 3D point-wise feature. For each 3D point-wise feature, DsCML can dynamically capture the related multiple 2D pixel-wise features with same category. This introduces much richer context information of texture and color in 2D image which is complementary to space information of 3D point cloud. Additionally, a novel sparse-to-dense learning loss is proposed to support the learning of multi-modality where 2D and 3D features differ by orders of magnitude. This DsCML is applied on source and target domain alternately which is the key to domain adaptation.

    The method mentioned above is adopted in an intra-domain manner where the 2D and 3D data contain same semantic content. In this paper, we further explore inter-domain cross modal learning for high-level semantic interaction of multi-modality data with different semantic content. Specifically, we introduce cross modal learning to common adversarial strategy by adding discriminator for identification between 2D and 3D features. We coin our method \textbf{C}ross \textbf{M}odal \textbf{A}dversarial \textbf{L}earning (CMAL). It enables the mutual learning between 2D and 3D as well as the alignment of feature distribution from different domains, which is the other key to domain adaptation.

    The main contributions of this paper are summarized as follows:
    \begin{itemize}
    \item To the best of our knowledge, this is the first work to explore cross modal learning in both intra and inter domain for the semantic segmentation problem.
    \item As for intra-domain cross modal learning, we propose a module named DsCML to establish sufficient relationships of multi-modality features, i.e., sparse-to-dense feature matching. 
    \item As for inter-domain cross modal learning, we propose a method named CMAL to achieve both high-level cross modal interaction and cross
    domain feature alignment. 
    \item The proposed method is evaluated on various real-to-real adaptation settings (i.e., day-to-night, country-to-country and dataset-to-dataset), obtaining state-of-the-art segmentation performance with both uni-modal and multi-modal methods.
    \end{itemize}


\section{Related Work}

    In this section, we briefly introduce the techniques related to our approach from three parts. We first give the description of Domain Adaptation in Sec. \ref{subsec:Unsupervised-Domain-Adaptation}. Besides, a considerable literature has grown up around the theme of Multi-Modality Learning in Sec. \ref{subsec:Multi-Modality-Learning}. Moreover, various relevant approaches about Adversarial Learning are discussed in Sec. \ref{subsec:Adversarial-learning}

    \begin{figure*}[t]
    \begin{centering}
    \includegraphics[scale=0.6]{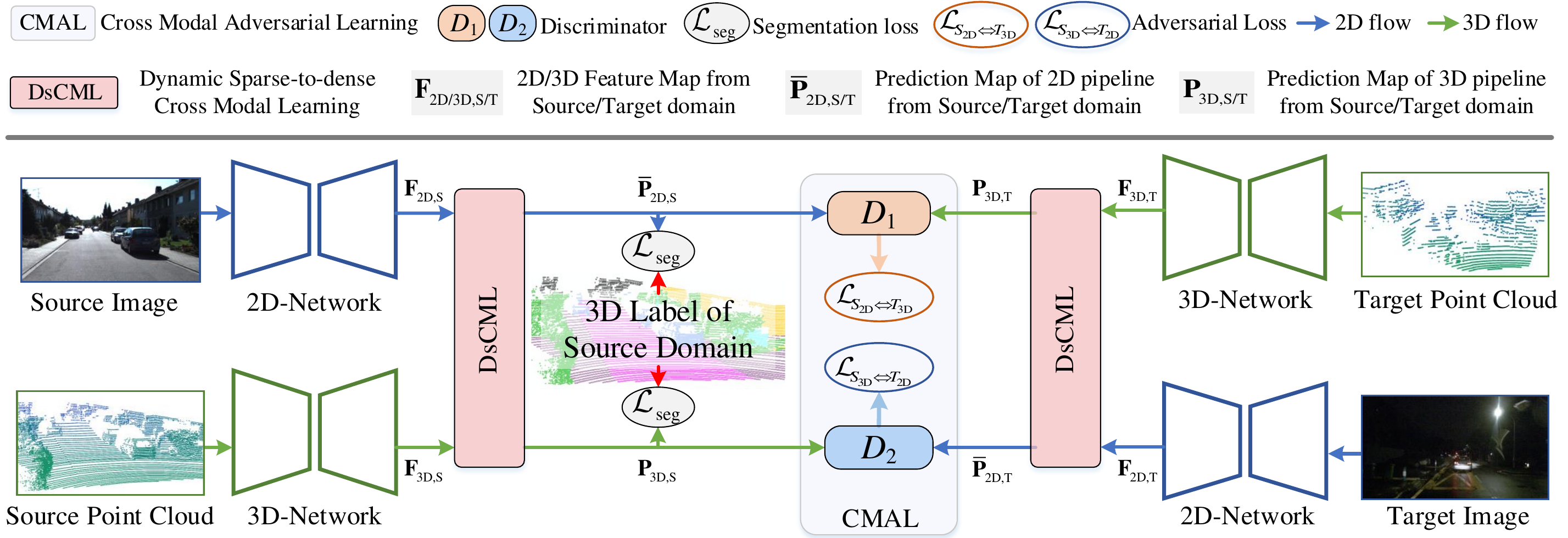} 
    \par\end{centering}
    \caption{Overall architecture of our approach, which consists of a DsCML module for intra-domain cross modal learning and one adversarial learning for inter-domain cross modal learning. At the end of 2D network and 3D network, DsCML can sufficiently transfer knowledge across multi-modality features (i.e., $\mathbf{F}_{2\mathrm{D},\mathrm{S/T}}$ and $\mathbf{F}_{3\mathrm{D},\mathrm{S/T}}$) and further generate the prediction maps for 3D semantic segmentation (i.e., $\bar{\mathbf{P}}_{2\mathrm{D,S/T}}$ and $\mathbf{P}_{3\mathrm{D,S/T}}$). Only the source predictions are supervised by 3D labels. After that, the prediction maps which are from different domains and different modalities are fed into CMAL for higher level cross modal learning.} \label{fig1} \vspace{-2mm}
    
    \end{figure*}


\subsection{Domain Adaptation\label{subsec:Unsupervised-Domain-Adaptation}}

    In the past few years, Domain adaptation has attracted great interest as its critical success in new, unseen environments. In the early years, several effective methods have been developed such as maximal confusion \cite{tzeng2015simultaneous,ganin2016domain,ganin2015unsupervised}, Maximum Mean Discrepancy (MMD) \cite{long2015learning,long2016unsupervised} and synthesizing images with target styles \cite{zhu2016generative,radford2015unsupervised}. Some other works advance adversarial training \cite{zhu2017unpaired,li2018semantic,hoffman2016fcns,chen2017no,hoffman2018cycada,tsai2018learning,peng2021overcoming,yang2020label} to narrow source-target distribution difference. Besides, as a typical semi-supervised learning scheme, self-training with pseudo-labels also show positive effect for domain adaptation \cite{zou2018unsupervised,zou2019confidence,li2019bidirectional,liu2020semi} and capture growing interest.

    While promising progress has been achieved, most algorithms focus on the single modality adaptation setting. It lacks the consideration of utilizing the complementary of multi-modality data. In addition, mostly methods are concerned with 2D semantic segmentation adaptation, few \cite{wu2019squeezesegv2} adopts domain adaptation in 3D segmentation from point clouds. In sight of this limitation, approach \cite{jaritz2020xmuda} on multi-modal input data (i.e., 2D image + 3D point cloud) has been proposed benefiting from multi-modal data. It assumes that both modalities are available on source and target domains. On this basis, in this paper, we aim to handle the problem in multi-modality domain adaption for 3D semantic segmentation.


\subsection{Multi-Modality Learning\label{subsec:Multi-Modality-Learning}}

    Taking the advantage of modality complementary is a straightforward and effective way to boost performance. A typical case is the fusion of RGB-Depth images for 2D semantic segmentation \cite{zhang2017amulet,valada2019self,hazirbas2016fusenet}. Due to both RGB and Depth images are with the almost same geometrical form, this kind of multi-modality learning is simple to implement. It is challenge to build the bridge for information interaction between 2D and 3D as the heterogeneous data form. A common solution is to filter dense 2D features to sparse point features to enable one-to-one 2D-3D feature matching which is convenient for subsequent processing \cite{ren2016faster,de2017fusion,de2018robust,kumar2020lidar,chiang2019unified,jaritz2019multi,su2018splatnet}. However, this type of feature matching leads to the lost of plentiful context information resulting in insufficient 2D-3D interaction. To this end, in this paper, we focus on how to exploit sparse-to-dense feature matching and corresponding learning strategy.


\subsection{Adversarial Learning\label{subsec:Adversarial-learning}}

    In domain adaptation, adversarial learning is mainly utilized to narrow the domain gap by reducing the distribution difference between source and target domain. Since the data representation are quite distinct among different feature levels in CNN, methods based on various feature spaces are presented. Among these methods, adaptation on pixel-level \cite{zhu2017unpaired,li2018semantic,hu2021learning,bousmalis2017unsupervised}, feature-level \cite{hoffman2016fcns,chen2017no,luo2019taking,hu2020randla}, both two levels \cite{hoffman2018cycada,vu2019advent}, output-space \cite{tsai2018learning} and label-space \cite{yang2020label} are exist.

    For instance, Chen \emph{et al}. \cite{chen2017no} implements a joint global and class-specific adversarial loss at the middle stage feature maps. Zhu \emph{et al}. \cite{zhu2017unpaired} addresses the adversarial learning on the pixel level, which essentially transfers the style of labeled source images into that of target domain. Hoffman \emph{et al}. \cite{hoffman2018cycada} proposes method CyCADA where both feature-level and pixel-level adversarial schemes are taken into account. Besides, Tsai \emph{et al.} \cite{tsai2018learning} has proven the effectiveness of output space feature alignment as it jointly promotes the optimization for both classifier and extractor. Moreover, a label-driven adversarial learning is studied in \cite{yang2020label} for semantic segmentation. In summary, adversarial learning is a high-level feature constraint towards the holistic data distribution. It enables the learning between two objects with different semantic content, so our method utilizes this advantage.


\section{Method}

    Our approach is presented for 3D semantic segmentation assuming the presence of 2D images and 3D point clouds. In this section, we first describe the architecture overview. Later, we showcase the intra-domain cross modal learning: DsCML in Section \ref{subsec:Intra-domain-cross-modal}. Finally, the inter-domain cross modal learning: CMAL is introduced in Section \ref{subsec:Inter-domain-cross-modal}.

    \begin{figure}[th]
    \centering{}\vspace{-2mm}
     \includegraphics[scale=0.17]{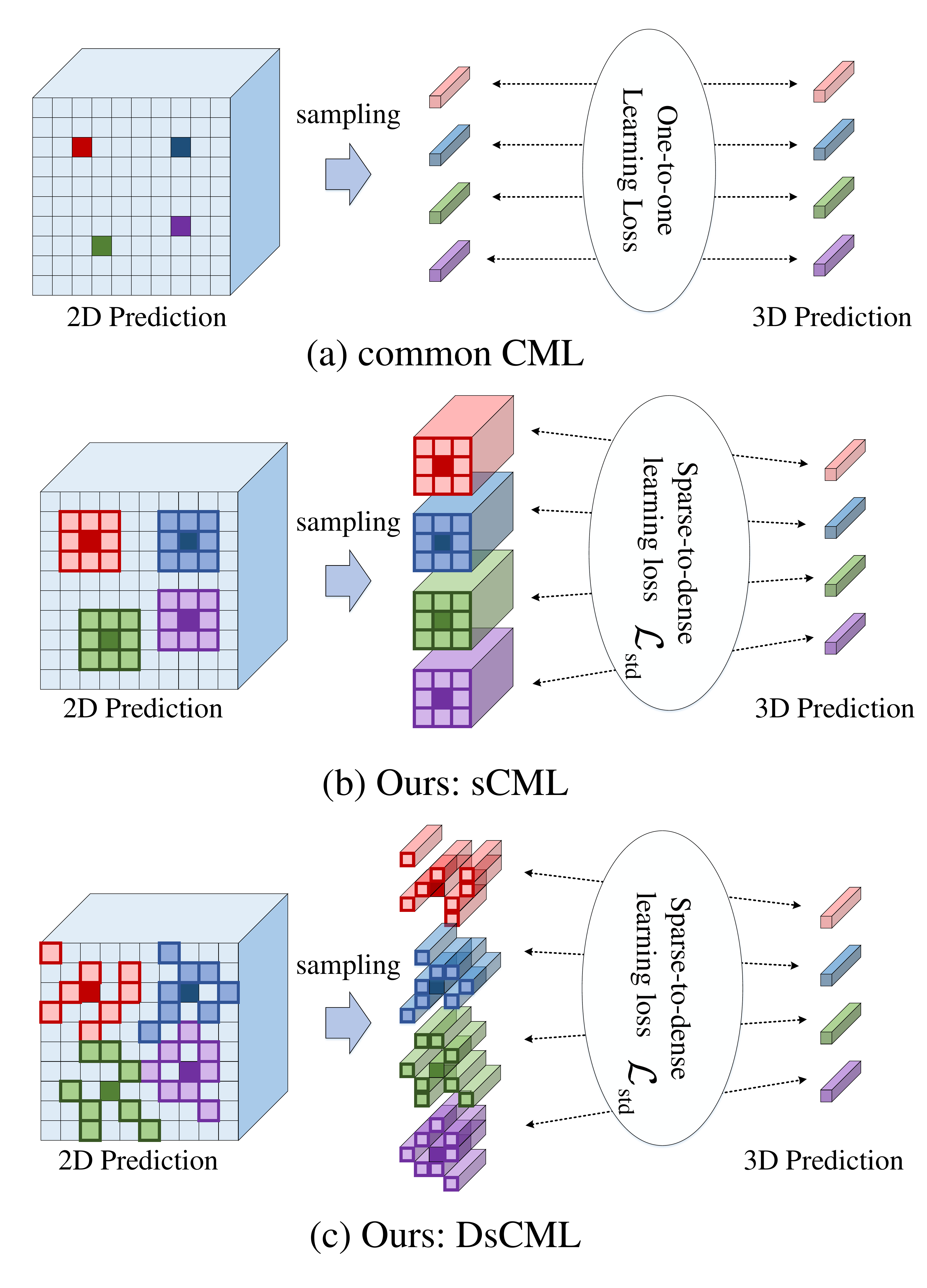} \caption{An illustration of feature matching in CML, sCML and DsCML. (a) In common CML, 3D point feature are one-to-one matched with the corresponding 2D pixel-wise features. (b) In sCML, each point features are matched with a square region learning with sparse-to-dense learning loss. (c) In DsCML, the region is deformable which enables the adaptive learning of model for searching the most suitable regions.}
    \label{--std-motivation}\vspace{-5mm}
     
    \end{figure}


\subsection{Overview of the Proposed Framework\label{subsec:Overview-of-the}}

    The overall architecture is shown in Fig. \ref{fig1} (best viewed from both sides to middle). We can briefly describe the main steps as follows. We begin by input the data of source domain $\mathrm{S}$ and target domain $\mathrm{T}$ into the 2D and 3D network to produce the feature maps before classifier (i.e., $\mathbf{F}_{2\mathrm{D},\mathrm{S}}$, $\mathbf{F}_{3\mathrm{D},\mathrm{S}}$, $\mathbf{F}_{2\mathrm{D},\mathrm{T}}$and $\mathbf{F}_{3\mathrm{D},\mathrm{T}}$). Next, the feature maps of each domain are fed into DsCML module for intra-domain cross modal learning. After that, DsCML generates the prediction for 3D semantic segmentation on both source and target domains (i.e., $\overline{\mathbf{P}}_{2\mathrm{D,S}}$, $\mathbf{P}_{3\mathrm{D,S}}$, $\overline{\mathbf{P}}_{2\mathrm{D,T}}$ and $\mathbf{P}_{3\mathrm{D,T}}$). It is worthy to mention that DsCML converts the dense 2D feature map into the prediction with the same size as 3D prediction. Hence, we use symbol with superscript (i.e., $\overline{\mathbf{P}}_{2\mathrm{D,S}}$ and $\overline{\mathbf{P}}_{2\mathrm{D,T}}$) to distinguish it from the 2D image segmentation prediction (i.e., $\mathbf{P}_{2\mathrm{D,S}}$ and $\mathbf{P}_{2\mathrm{D,T}}$). Afterwards, only the source predictions are supervised by the label of source domain. Finally, the source 2D (3D) and target 3D (2D) predictions are fed into the discriminator $D_{1}$ ($D_{2}$) for adversarial learning, aiming to execute intra-domain cross modal learning. 

    \begin{figure*}[t]
    \begin{centering}
    \includegraphics[scale=0.48]{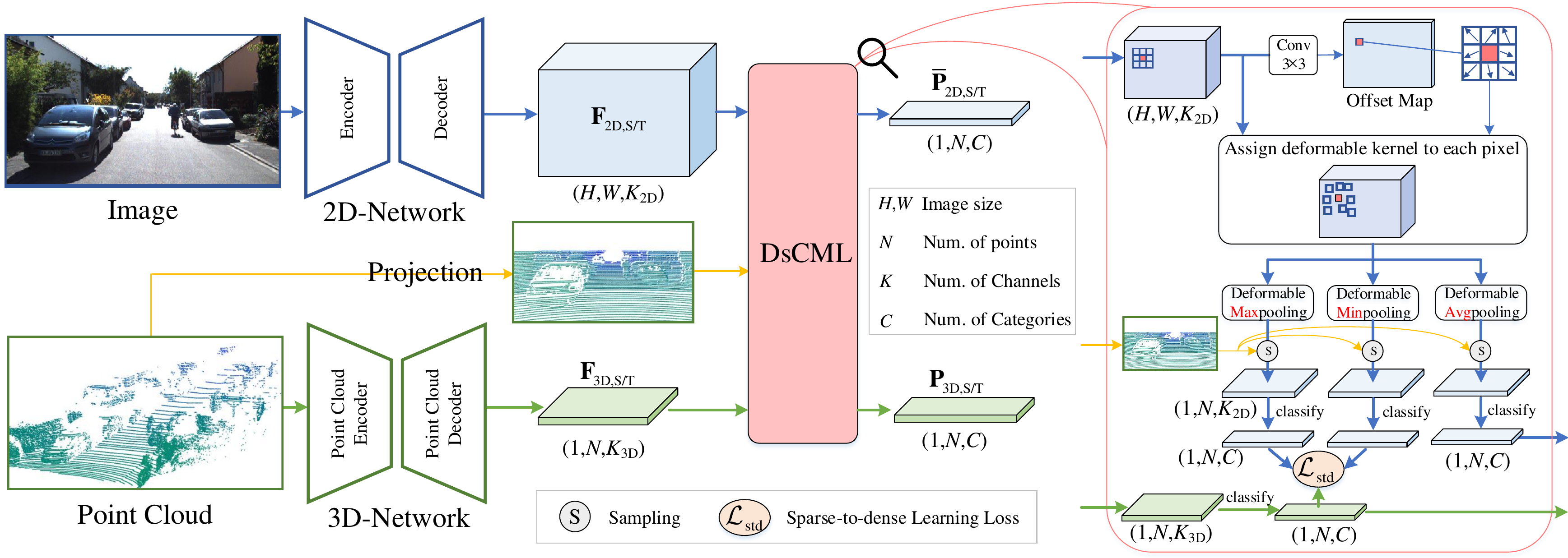} 
    \par\end{centering}
    \caption{The architecture of our proposed DsCML. With the help of the deformable pooling \cite{dai2017deformable} accompanied with the proposed $\mathcal{L}_{std}$, we can achieve the dynamic sparse-to-dense feature matching.}
    \label{fig-DsCML} \vspace{-2mm}
    
    \end{figure*}


\subsection{Intra-domain cross modal learning: DsCML\label{subsec:Intra-domain-cross-modal}}

    The goal of DsCML is to enable sufficient information interaction between the modalities in an intra-domain manner to make them complement each other. This objective can be applied to target domain as its training without access of any annotations. The cross modal learning on target domain enables the domain adaptation for 3D semantic segmentation.

    \textbf{Cross Modal Learning (CML).} To better understand how DsCML solves the problem of insufficient cross modal learning, we begin with a common Cross Modal Learning method (CML). As shown in Fig. \ref{--std-motivation} (a), the loss is implemented on each matched feature pair obtained by sparse-to-sparse matching. Note that we choose 2D and 3D feature maps from the output space where features are already handled by classifier. Hence, it can be formulated as:

    \begin{equation}
    \mathcal{L}=\frac{1}{N}\sum_{n=1}^{N}\mathcal{K}(\mathrm{Samp}(\mathbf{P}_{2\mathrm{D}})^{n},\mathbf{P}_{3\mathrm{D}}^{n}),
    \end{equation}
    where $\mathrm{Samp}(\mathbf{P}_{2\mathrm{D}})^{n}$ denotes the probability scores of the $n$-th sampled pixel, $\mathbf{P}_{3\mathrm{D}}^{n}$ is probability scores of the $n$-th point, $N$ denotes the number of points in 3D point cloud and $\mathcal{K}(\cdot,\cdot)$ represents the KL distance. We can see that in common CML, only part of the features in 2D feature map are learning with 3D point features, large amount of useful features are discarded.

    \textbf{Sparse-to-dense Cross Modal Learning (sCML).} As is well-known, the neighboring pixels mostly belong to a same category in 2D semantic segmentation. In light of this, we try to utilize a square patch of 2D feature map to exchange information with the corresponding point features of 3D point cloud. As shown in Fig. \ref{--std-motivation} (b), the sparse-to-dense learning loss is the key to implement one-to-many constraints. It can be written as:

    \begin{equation}
    \mathcal{L}_{std}=\frac{1}{N}\sum_{n=1}^{N}\mathcal{K}(\phi_{max}^{n}(\mathbf{P}_{2\mathrm{D}}),\mathbf{P}_{3\mathrm{D}}^{n})+\mathcal{K}(\phi_{min}^{n}(\mathbf{P}_{2\mathrm{D}}),\mathbf{P}_{3\mathrm{D}}^{n}),
    \end{equation}
    where $\phi_{max}^{n}(\mathbf{P}_{\mathrm{2D}})$ denotes the max probability scores in the $n$-th 2D patch and similarly $\phi_{min}^{n}(\mathbf{P}_{\mathrm{2D}})$ is the min probability scores in the $n$-th 2D patch. By constraining the supremum and infimum of a set, we can progressively constrain every element in the patch after several iterations.

    \textbf{Dynamic sparse-to-dense Cross Modal Learning (DsCML).} As shown in Fig. \ref{--std-motivation} (c), we further introduce deformable patch to adaptively search the patch with appropriate region. The sparse-to-dense loss $\mathcal{L}_{std}$ can thus be improved as:

    \begin{equation}
    \mathcal{L}_{std}=\frac{1}{N}\sum_{n=1}^{N}\mathcal{K}(\Phi_{max}^{n}(\mathbf{P}_{2\mathrm{D}}),\mathbf{P}_{3\mathrm{D}}^{n})+\mathcal{K}(\Phi_{min}^{n}(\mathbf{P}_{2\mathrm{D}}),\mathbf{P}_{3\mathrm{D}}^{n}),\label{eq:std}
    \end{equation}
    where $\Phi_{max}^{n}(\mathbf{P}_{\mathrm{2D}})$ denotes the max probability scores in the $n$-th 2D deformable patch and $\Phi_{min}^{n}(\mathbf{P}_{\mathrm{2D}})$ is the min one.

    \textbf{Architecture of DsCML.} Fig. \ref{fig-DsCML} illustrates the architecture of our DsCML module for intra-domain cross modal learning. Borrowing from the Deformable Convolution \cite{dai2017deformable}, we utilize the extracted offset map to enable the deformable max/min/avg pooling. After pooling, we can select the max/min/avg features of each deformable patch with the guidance of projection from 3D to 2D. Finally, the max/min/avg probability scores are obtained through the classifier. The max and min probability scores are utilized for cross modal learning in DsCML, while the avg scores are fed into the computation of segmentation loss:

    \begin{equation}
    \mathcal{L}_{seg}=-\frac{1}{N}\sum_{n=1}^{N}Y_{\mathrm{S}}^{n}(\log(\Phi_{avg}^{n}(\mathbf{P}_{2\mathrm{D,S}}))+\log(\mathbf{P}_{3\mathrm{D,S}}^{n})),
    \end{equation}
    where $Y_{S}^{n}$ denotes the source label of the $n$-th point and $\Phi_{avg}^{n}(\mathbf{P}_{2\mathrm{D,S}})$ is the average probability scores in the $n$-th deformable patch of source 2D predicted map.


\subsection{Inter-domain cross modal learning: CMAL\label{subsec:Inter-domain-cross-modal}}

    Although performing intra-domain cross modal learning can directly improve performance of each modal, there still exist two problems: (a) the learning is mainly towards different modal features with same semantic content, which is a low-level feature alignment; (b) the learning on source and target domain are mutually independent, resulting in the supervision of source label cannot effectively guide the segmentation of target domain. For the latter, Adversarial Learning has shown tremendous adaptation progress by focusing on the mapping between data from different domains in feature space. In light of this, we advance to apply Cross Modal Adversarial Learning (CMAL) to solve both problems with one scheme. In CMAL, the multi-modality learning is execute towards features with different content, different modal and different domain. This high-level feature alignment aims to relieve the distribution difference between source and target as well as 2D and 3D. In this way, the 2D and 3D networks can jointly act 3D semantic segmentation on both domains.

    As for CMAL, let $\overline{\mathbf{P}}_{\mathrm{2D,S}},\mathbf{P}_{\mathrm{3D,S}}=G(x_{\mathrm{2D,S}},x_{\mathrm{3D,S}})$ be the 3D prediction maps of source 2D image $x_{\mathrm{2D,S}}$ and 3D point cloud $x_{\mathrm{3D,S}}$, where G denotes the multi-modality network with DsCML. Similarly, we process the data in target domain and obtain the predictions $\overline{\mathbf{P}}_{\mathrm{2D,T}}$ and $\mathbf{P}_{\mathrm{3D,T}}$.

    To make the distribution of $\overline{\mathbf{P}}_{\mathrm{2D,S}}$ ($\mathbf{P}_{\mathrm{3D,S}}$) closer to $\mathbf{P}_{\mathrm{3D,T}}$ ($\overline{\mathbf{P}}_{\mathrm{2D,T}}$), we use adversarial loss as:

    \vspace{-5 mm}
    \begin{equation}
    \mathcal{L}_{S_{\mathrm{2D}}\Leftrightarrow T_{\mathrm{3D}}}=-\log(\rho(\overline{\mathbf{P}}_{\mathrm{2D,S}}))-\log(1-\rho(\mathbf{P}_{3\mathrm{D,T}})),
    \end{equation}
    \vspace{-8 mm}
    
    \begin{equation}
    \mathcal{L}_{S_{\mathrm{3D}}\Leftrightarrow T_{\mathrm{2D}}}=-\log(\rho(\mathbf{P}_{\mathrm{3D,S}}))-\log(1-\rho(\overline{\mathbf{P}}_{2\mathrm{D,T}})),
    \end{equation}
    where $\rho(\mathbf{P})$ be the probability that the prediction map $\mathbf{P}$ belongs to the source domain after the identification of discriminator.

    We optimize the following min-max criterion:

    \begin{equation}
    \underset{G}{\mathrm{max}}\underset{D_{1}}{\mathrm{min}}\mathcal{L}_{S_{\mathrm{2D}}\Leftrightarrow T_{\mathrm{3D}}},\label{eq:maxmin1}
    \end{equation}
    
    \begin{equation}
    \underset{G}{\mathrm{max}}\underset{D_{2}}{\mathrm{min}}\mathcal{L}_{S_{\mathrm{3D}}\Leftrightarrow T_{\mathrm{2D}}},\label{eq:maxmin2}
    \end{equation}
    where $D_{1}$ and $D_{2}$ are two discriminators with same architecture to handle different adversarial tasks. We handle Eq. \ref{eq:maxmin1} and \ref{eq:maxmin2} by alternating optimization between $G$ and $D_{1}$ ($D_{2}$).

    \begin{table}
    \caption{The sample number in each split of datasets for all three settings. Note that the training samples in target domain are without labels.
    \label{tab:Size-of-the}}
    \vspace{0.2cm}
    \centering{}\doublerulesep=0.5pt \resizebox{0.45\textwidth}{!}{%
    \begin{tabular}{cccccc}
    \hline 
    \multirow{2}{*}{settings} & Source  &  & \multicolumn{3}{c}{Target}\tabularnewline
    \cline{2-2} \cline{4-6}
     & train  &  & train  & val  & test\tabularnewline
    \hline 
    nuScenes:Day/Night  & 24745  &  & 2779  & 606  & 602\tabularnewline
    nuScenes:USA/Singapore  & 15695  &  & 9665  & 2770  & 2929\tabularnewline
    A2D2/SemanticKITTI  & 27695  &  & 18029  & 1101  & 4071\tabularnewline
    \hline 
    \end{tabular}}
    \vspace{-0.5cm}
    \end{table}


    \begin{table*}[t]
    \caption{Comparison results with both uni-modal and multi-modal adaptation methods for 3D semantic segmentation in different cross-modal domain adaptation settings. We report the result for each network stream in terms of mIoU. The best two results are marked in bold and underline. \textquoteleft Avg\textquoteright{}  denotes the result which is obtained by taking the mean of the predicted 2D and 3D probabilities after softmax.\label{tab:experiment-results}}
    \vspace{0.3cm}
    \centering{}\doublerulesep=0.5pt \resizebox{1.0\textwidth}{!}{%
    \begin{tabular}{cl>{\centering}p{1.2cm}>{\centering}p{1.2cm}>{\centering}p{1.2cm}>{\centering}p{0.3cm}>{\centering}p{1cm}>{\centering}p{1cm}>{\centering}p{1cm}>{\centering}p{0.3cm}>{\centering}p{1.2cm}>{\centering}p{1.2cm}>{\centering}p{1.2cm}>{\centering}p{0.3cm}>{\centering}p{1cm}>{\centering}p{1cm}>{\centering}p{1cm}>{\centering}p{0.3cm}>{\centering}p{1.2cm}>{\centering}p{1.2cm}>{\centering}p{1.2cm}}
    \hline 
    \multirow{2}{*}{Modality} & \multirow{2}{*}{~~~~~~~~~~~~Method} & \multicolumn{3}{c}{USA$\rightarrow$Singapore (nuScenes)} &  & \multicolumn{3}{c}{Day$\rightarrow$Night (nuScenes)} &  & \multicolumn{3}{c}{USA$\rightarrow$Singapore (Lidarseg)} &  & \multicolumn{3}{c}{Day$\rightarrow$Night (Lidarseg)} &  & \multicolumn{3}{c}{A2D2$\rightarrow$SemanticKITTI}\tabularnewline
    \cline{3-5} \cline{7-9} \cline{11-13} \cline{15-17} \cline{19-21} 
     &  & 2D & 3D & Avg &  & 2D & 3D & Avg &  & 2D & 3D & Avg &  & 2D & 3D & Avg &  & 2D & 3D & Avg\tabularnewline
    \hline 
     & Baseline (Source only) & 53.2 & 46.8 & 61.2 &  & 41.8 & 41.4 & 47.6 &  & 53.3 & 48.0 & 61.6 &  & 41.8 & 43.8 & 48.0 &  & 36.4 & 37.3 & 42.2\tabularnewline
    \hline 
    \multirow{6}{*}{Uni-modal} & MinEnt \cite{vu2019advent} & 53.4 & 47.0 & 59.7 &  & 44.9 & 43.5 & 51.3 &  & 53.6 & 48.6 & 61.9 &  & 44.9 & 44.3 & 51.8 &  & 38.8 & 38.0 & 42.7\tabularnewline
     & PL \cite{li2019bidirectional} & 55.5 & 51.8 & 61.5 &  & 43.7 & 45.1 & 48.6 &  & 55.4 & 52.7 & 62.8 &  & 43.9 & 47.6 & 50.9 &  & 37.4 & 44.8 & 47.7\tabularnewline
     & FCNs in the Wild \cite{hoffman2016fcns} & 53.7 & 46.8 & 61.0 &  & 42.6 & 42.3 & 47.9 &  & 54.0 & 49.2 & 62.4 &  & 42.6 & 43.9 & 48.7 &  & 37.1 & 43.5 & 43.6\tabularnewline
     & CyCADA \cite{hoffman2018cycada} & 54.9 & 48.7 & 61.4 &  & 45.7 & 45.2 & 49.7 &  & 54.9 & 51.3 & 62.6 &  & 45.5 & 47.8 & 49.6 &  & 38.2 & 43.9 & 43.9\tabularnewline
     & AdaptSegNet \cite{tsai2018learning} & 56.3 & 47.7 & 61.8 &  & 45.3 & 44.6 & 49.6 &  & 56.5 & 49.0 & 62.0 &  & 45.5 & 45.3 & 49.3 &  & 38.8 & 44.3 & 44.2\tabularnewline
     & CLAN \cite{luo2019taking} & 57.8 & 51.2 & 62.5 &  & 45.6 & 43.7 & 49.2 &  & 57.7 & 52.1 & 63.1 &  & 45.6 & 45.1 & 50.1 &  & 39.2 & 44.7 & 44.5\tabularnewline
    \hline 
    \multirow{5}{*}{Multi-modal} & xMUDA \cite{jaritz2020xmuda} & 59.3 & 52.0 & 62.7 &  & 46.2 & 44.2 & 50.0 &  & 61.7 & 52.6 & 63.3 &  & 47.3 & 46.0 & 50.6 &  & 36.8 & 43.3 & 42.9\tabularnewline
     & xMUDA+PL \cite{jaritz2020xmuda} & 61.1 & 54.1 & 63.2 &  & 47.1 & 46.7 & 50.8 &  & 63.0 & 54.3 & 64.2 &  & 48.4 & 47.5 & 51.2 &  & 43.7 & 48.5 & 49.1\tabularnewline
    \cline{2-21} 
     & DsCML & 61.3 & 53.3 & 63.6 &  & 48.0 & 45.7 & 51.0 &  & 63.3 & 54.0 & 64.2 &  & 49.8 & 47.2 & 51.7 &  & 39.6 & 45.1 & 44.5\tabularnewline
     & DsCML + CMAL & \uline{63.4} & \uline{55.6} & \uline{64.8} &  & \uline{49.5} & \uline{48.2} & \uline{52.7} &  & \textbf{65.6} & \uline{56.2} & \uline{66.1} &  & \uline{50.9} & \uline{49.3} & \uline{53.2} &  & \uline{46.3} & \uline{50.7} & \uline{51.0}\tabularnewline
     & DsCML + CMAL + PL & \textbf{63.9} & \textbf{56.3} & \textbf{65.1} &  & \textbf{50.1} & \textbf{48.7} & \textbf{53.0} &  & \textbf{65.6} & \textbf{57.5} & \textbf{66.9} &  & \textbf{51.4} & \textbf{49.8} & \textbf{53.8} &  & \textbf{46.8} & \textbf{51.8} & \textbf{52.4}\tabularnewline
    \hline 
    \end{tabular}}
    \vspace{-0.2cm}
    \end{table*}


\section{Experiments}


\subsection{Datasets Description}

    We strictly follow Jaritz's work: xMUDA \cite{jaritz2020xmuda} to implement our method on three real-to-real adaptation settings: day-to-night, country-to-country and dataset-to-dataset. Three public datasets nuScenes \cite{caesar2020nuscenes}, A2D2 \cite{geyer2020a2d2} and SemanticKITTI \cite{behley2019semantickitti} are leveraged where the LiDAR and camera are synchronized and calibrated. Only 3D annotations are utilized for 3D semantic segmentation. Specifically, we leverage nuScenes to generate the splits: Day/Night and USA/Singapore for day-to-night and country-to-country adaptation. The other two datasets are utilized for dataset-to-dataset adaptation, i.e, A2D2/SemanticKITTI. Tab. \ref{tab:Size-of-the} shows the split details of three datasets for three real-to-real adaptation settings.

    As for Day/Night, the 3D point clouds captured by LiDAR show small domain difference due to the sensor has a strong robustness to light variations. While the RGB image capture by camera is the opposite. In the setting of USA/Singapore, the 3D domain difference may be larger than that of 2D in some conditions or vice versa. In A2D2/SemanticKITTI, the density (resolution) of point cloud are large different which highly affect the adaptation performance of 3D network. In this case, the image with small domain gap can help to boost the adaptation performance.


\subsection{Implementation Details}

    \textbf{Dataset Preprocessing:} For domain adaptations of Day/Night and USA/Singapore in nuScenes \cite{caesar2020nuscenes}, we utilize the accessible 3D bounding boxes annotations to obtain the 3D point-wise labels as xMUDA \cite{jaritz2020xmuda} did. More specifically, for the point lying inside 3D boxes, we assign it the corresponding object label, otherwise it is labeled as background. To make a more convincing evaluation on our approach, we also experiment on nuScenes-Lidarseg \cite{caesar2020nuscenes} (shorten to \textquoteleft Lidarseg\textquoteright) which contains the point-wise annotation. For domain adaptation of A2D2\cite{geyer2020a2d2}/SemanticKITTI\cite{behley2019semantickitti}, we select 10 classes which are shared between the two datasets, i.e., Car, Truck, Bike, Person, Road, Sidewalk, Parking, Nature, Building and Other objects. With the help of code released from xMUDA \cite{jaritz2020xmuda}, we compute the projection between each 3D point and its corresponding 2D image pixel.

    \textbf{Network Baseline:} To make a fair comparison with the only known multi-modal 3D domain adaptation method \cite{jaritz2020xmuda}, for 2D network, we adopt ResNet34 \cite{he2016deep} pre-trained on ImageNet \cite{deng2009imagenet} as the encoder of U-Net \cite{ronneberger2015u}. For 3D network, we use SparseConvNet \cite{graham20183d} with U-Net architecture and implement downsampling for six times. Meanwhile, a voxel with size of 5cm is adopted in 3D network, which is small enough to ensure only one 3D point exists in each voxel. The source codes and models are trained and evaluated on PyTorch toolbox \cite{paszke2017automatic} based on Python 3.7 platform. All proposed models are implemented on one NVIDIA RTX 3090Ti GPU with 24GB RAM and four E-2224 CPUs.
    
    \textbf{Parameter Settings:} In training period, we choose a batch size of 8 and Adaptive Moment Estimation (Adam) \cite{kingma2014adam} optimizer with $\beta_{1}=0.9$ and $\beta_{2}=0.999$. The learning rate is set to $1e^{-3}$ initially and follows the poly learning rate policy \cite{chen2017rethinking} with a poly power of 0.9. Each deformable patch in DsCML is based on $5\times5$ square patch. The max training iteration is set to 100k.

    \textbf{Evaluation:} Following previous domain adaptation works, we evaluate the performance of a model on the test set by using the standard PASCAL VOC intersection-overunion (IoU). The mean IoU (mIoU) is the mean of all IoU values over all categories. Specifically, the mIoU can be written as follows:

    \begin{equation}
    \mathrm{mIoU}=\frac{1}{C}\sum_{i=0}^{C}\frac{TP(i)}{TP(i)+FP(i)+FN(i)},
    \end{equation}
    where $C$ is the overall number of categories, $TP(i)$, $FP(i)$, and $FN(i)$ are values of true positive, false positive and false negative towards the $i$-th category, respectively.


\subsection{Comparative Studies}

    We evaluate our approach on the above three real-to-real adaptation settings and compare with some representative uni-modal domain adaption methods: MinEnt \cite{vu2019advent}, pseudo-labeling (PL) \cite{li2019bidirectional}, FCNs in the Wild \cite{hoffman2016fcns}, CyCADA \cite{hoffman2018cycada}, AdaptSegNet \cite{tsai2018learning} and CLAN \cite{luo2019taking}. These uni-modal domain adaptation methods are evaluated on each modality with the same network baselines as ours, i.e., U-Net with ResNet34 encoder (2D network) and SparseConvNet (3D network). In the output space of 2D pipeline, we sample the features from the feature map outputted by 2D network according to the projection from 3D to 2D. Since AdaptSegNet \cite{tsai2018learning} adopts adversarial learning in the output space, regarding the 2D pipeline of AdaptSegNet, we are faced with two options: implementing adversarial learning on 2D feature map or sampled point features. Herein, we choose the one with better performance from two options, i.e., adaption on sampled point features. Besides, we compare our approach with the only known multi-modal domain adaption method: xMUDA \cite{jaritz2020xmuda}.

\begin{figure*}[t]
    \begin{centering}
    \includegraphics[scale=0.2]{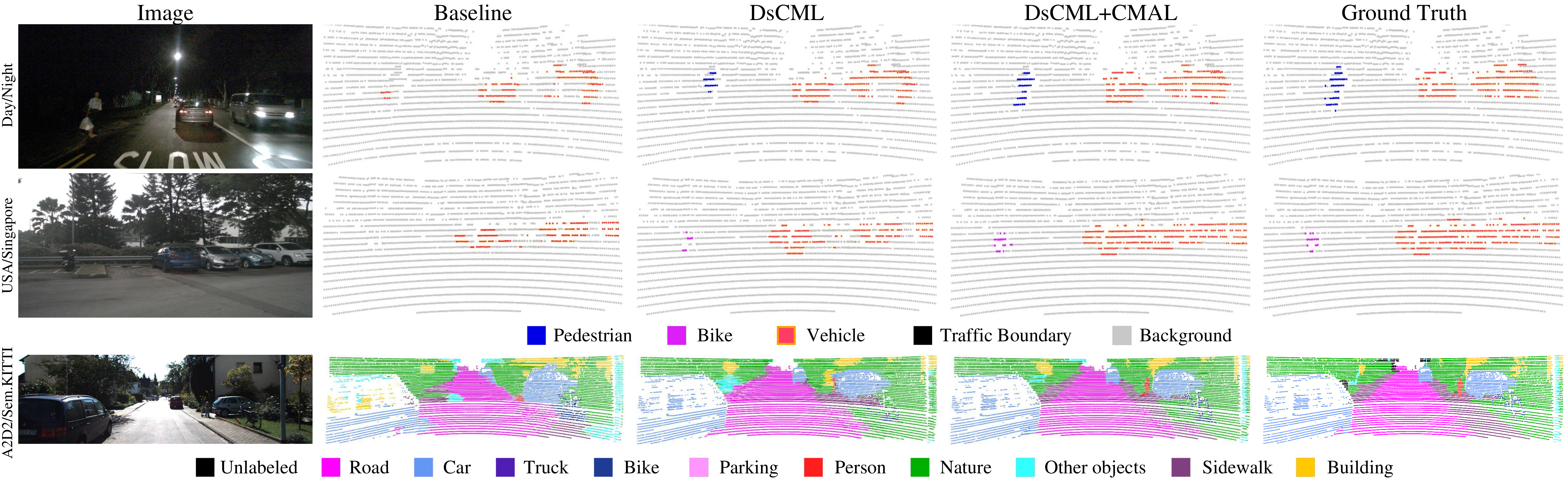}
    \par\end{centering}
    \caption{Qualitative 3D semantic segmentation results on three multi-modality adaptation settings: Day/Night, USA/Singapore and A2D2/Sem.KITTI. We show the ensembling results by averaging the softmax outputs of 2D and 3D networks. It can be seen that the segmentation performance is improved by adding the proposed modules progressively.\label{fig:visualization}}
    \vspace{0mm}
    \end{figure*}

    All comparison results for 3D semantic segmentation are reported in Tab. \ref{tab:experiment-results}. We can observe that the only usage of DsCML and CMAL brings a significant adaptation effect on all settings compared to Baseline (source only). It is worth noting that our model with only DsCML can outperform all state-of-the-art uni-modal methods. It proves that the two modalities (2D and 3D) are indeed complementary to each other and our DsCML can consistently improves performance of both modalities. From the comparison with ``xMUDA+PL'' which is the final approah in \cite{jaritz2020xmuda}, our model with both DsCML and CMAL achieves the superior performance and contributes 2.5\% (2D), 1.8\% (3D) and 1.9\% (Avg) mIoU gains on average over all settings. Moreover, the model only with DsCML also outperforms ``xMUDA'' by 2.1\% (2D), 1.4\% (3D) and 1.1\% (Avg) on average. Some qualitative segmentation examples can be viewed in Fig. \ref{fig:visualization}.


\subsection{Ablation Studies}


\subsubsection{Effects of DsCML and CMAL}

    Next, we conduct additional experiments to demonstrate the benefits of our proposed methods. In Tab. \ref{tab:experiment-results}, we detail the performance of each design as well as its mIoU improvement by progressively adding DsCML (intra-domain cross modal learning) and CMAL (inter-domain cross modal learning) from the baseline. We can see that DsCML helps to significantly boots the performance of 2D network by 3.2\%$\sim$10\%, 3D network by 3.4\%$\sim$7.8\%, Avg by 2.3\%$\sim$3.7\%. With the addition of CMAL, our model achieves over 1.1\% improvement on 2D network and over 2.1\% improvement on 3D network. This confirms the effectiveness of our DsCML and CMAL. Besides, we also conduct experiments with PL to demonstrate the complementarity of our model and PL.


\subsubsection{Effects of sparse-to-dense feature matching and deformable patch in DsCML}

    \begin{table}
    \caption{Performance comparison on CML, sCML and DsCML. Each improvement is obtained by comparing with the upper model. \label{tab:ablation-stdfeam-deform}}
    \vspace{0.3cm}
    \centering{}\doublerulesep=0.5pt \resizebox{0.48\textwidth}{!}{%
    \begin{tabular}{ll>{\centering}p{1.1cm}>{\centering}p{1.1cm}>{\centering}p{1.1cm}>{\centering}p{1.1cm}>{\centering}p{1.1cm}>{\centering}p{1.1cm}>{\centering}p{0.9cm}>{\centering}p{0.9cm}>{\centering}p{0.9cm}}
    \hline 
    \multirow{2}{*}{} & \multirow{2}{*}{Method} & \multicolumn{3}{c}{USA$\rightarrow$Singapore (nuScenes)} & \multicolumn{3}{c}{Day$\rightarrow$Night (nuScenes)} & \multicolumn{3}{c}{A2D2$\rightarrow$Sem.KITTI}\tabularnewline
    \cline{3-11} 
     &  & 2D  & 3D & Avg & 2D  & 3D & Avg & 2D  & 3D & Avg\tabularnewline
    \hline 
    (a)  & CML  & 59.6  & 51.7  & 62.4 & 46.3  & 44.3  & 49.8 & 36.4  & 42.8 & 42.3\tabularnewline
    \hline 
    \multirow{2}{*}{(b)} & \multirow{2}{*}{sCML} & 60.6  & 52.5  & 63.2 & 47.2  & 45.1  & 50.7 & 38.2  & 44.3 & 44.0\tabularnewline
     &  & ($\uparrow$1.0)  & ($\uparrow$0.8) & ($\uparrow$0.8) & ($\uparrow$0.9)  & ($\uparrow$0.8) & ($\uparrow$0.9) & ($\uparrow$1.8)  & ($\uparrow$1.5) & ($\uparrow$0.7)\tabularnewline
    \hline 
    \multirow{2}{*}{(c)} & \multirow{2}{*}{DsCML} & 61.3  & 53.3  & 63.6 & 48.0  & 45.7  & 51.0 & 39.6  & 45.1 & 44.5\tabularnewline
     &  & ($\uparrow$0.7)  & ($\uparrow$0.8)  & ($\uparrow$0.4) & ($\uparrow$0.8)  & ($\uparrow$0.6)  & ($\uparrow$0.3) & ($\uparrow$1.4)  & ($\uparrow$0.8) & ($\uparrow$0.5)\tabularnewline
    \hline 
    \end{tabular}} 
    \vspace{-0.2cm}
    \end{table}

    To evaluate the benefits of the proposed sparse-to-dense feature matching in DsCML, we re-implement our approach with CML and sCML, respectively. The comparison results are reported in Tab. \ref{tab:ablation-stdfeam-deform}. Compared to CML (Tab. \ref{tab:ablation-stdfeam-deform} a), sCML (Tab. \ref{tab:ablation-stdfeam-deform} b) can stably boost the results over 0.7\% on each modal under three settings. It demonstrates the benefits of sparse-to-dense feature matching. To demonstrate the effect of deformable patch in DsCML, we also conducted an ablation experiment by comparing our DsCML with sCML. According to the comparison between model only sCML (Tab. \ref{tab:ablation-stdfeam-deform} b) and model only with DsCML (Tab. \ref{tab:ablation-stdfeam-deform} c), we can see that the proposed method DsCML achieves stable improvements and performs best on all three settings. It means the deformable strategy can find the suitable region of patch to exchange information with 3D point features.

\subsubsection{Effects of cross modal alignment in CMAL}

    As shown in Tab. \ref{tab:Ablation-studies-on}, we conduct ablation studies on three options: (a) inter-modal alignment, (b) cross-modal alignment and (c) both. Except two scores slightly falling behind the best, (b) shows the best performances in all settings. Compared with (a) which only considers relieving domain gap, (b) can effectively improve adaptation performances by simultaneously narrow the domain and modality gap. Although (c) adopts both schemes, it shows unstable performances as it introduces much more discriminators leading to training complexity and difficulty.
    
    \begin{table}[t]
    \caption{Ablation on feature alignment styles in CMAL. The best two results are marked in bold and underline. \label{tab:Ablation-studies-on}}
    \vspace{0.3cm}
    \centering{} \resizebox{0.48\textwidth}{!}{%
    \begin{tabular}{ccccccccccc}
    \hline
    \multirow{2}{*}{} & \multirow{2}{*}{Adversary Options} & \multicolumn{3}{c}{U$\rightarrow$S (Lidarseg)} & \multicolumn{3}{c}{D$\rightarrow$N (Lidarseg)} & \multicolumn{3}{c}{A2D2$\rightarrow$Sem.KITTI}\tabularnewline
    \cline{3-11}
     &  & 2D & 3D & Avg & 2D & 3D & Avg & 2D & 3D & Avg\tabularnewline
    \hline 
    (a) & $\mathrm{S_{2D}}\Leftrightarrow\mathrm{T_{2D}}$ \& $\mathrm{S_{3D}}\Leftrightarrow\mathrm{T_{3D}}$ & \underline{64.8} & 54.9 & \underline{64.7} & \underline{50.1} & 47.3 & 52.5 & 40.1 & \underline{48.2} & 48.4\tabularnewline
    (b) & $\mathrm{S_{2D}}\Leftrightarrow\mathrm{T_{3D}}$ \& $\mathrm{S_{3D}}\Leftrightarrow\mathrm{T_{2D}}$ & \textbf{65.6} & \textbf{56.2} & \textbf{66.1} & \textbf{50.9} & \underline{49.3} & \textbf{53.2} & \underline{46.3} & \textbf{50.7} & \textbf{51.0}\tabularnewline
    (c) & Both (a) and (b) & 63.8 & \underline{55.3} & 64.4 & 49.6 & \textbf{49.7} & \underline{52.9} & \textbf{46.6} & 47.5 & \underline{49.2}\tabularnewline
    \hline 
    \end{tabular}}
    \vspace{-0.5cm}
    \end{table}


\subsubsection{Effects of Sparse-to-dense Loss}

\begin{figure}[t]
    \vspace{0cm}
    \centering{}
     \includegraphics[scale=0.58]{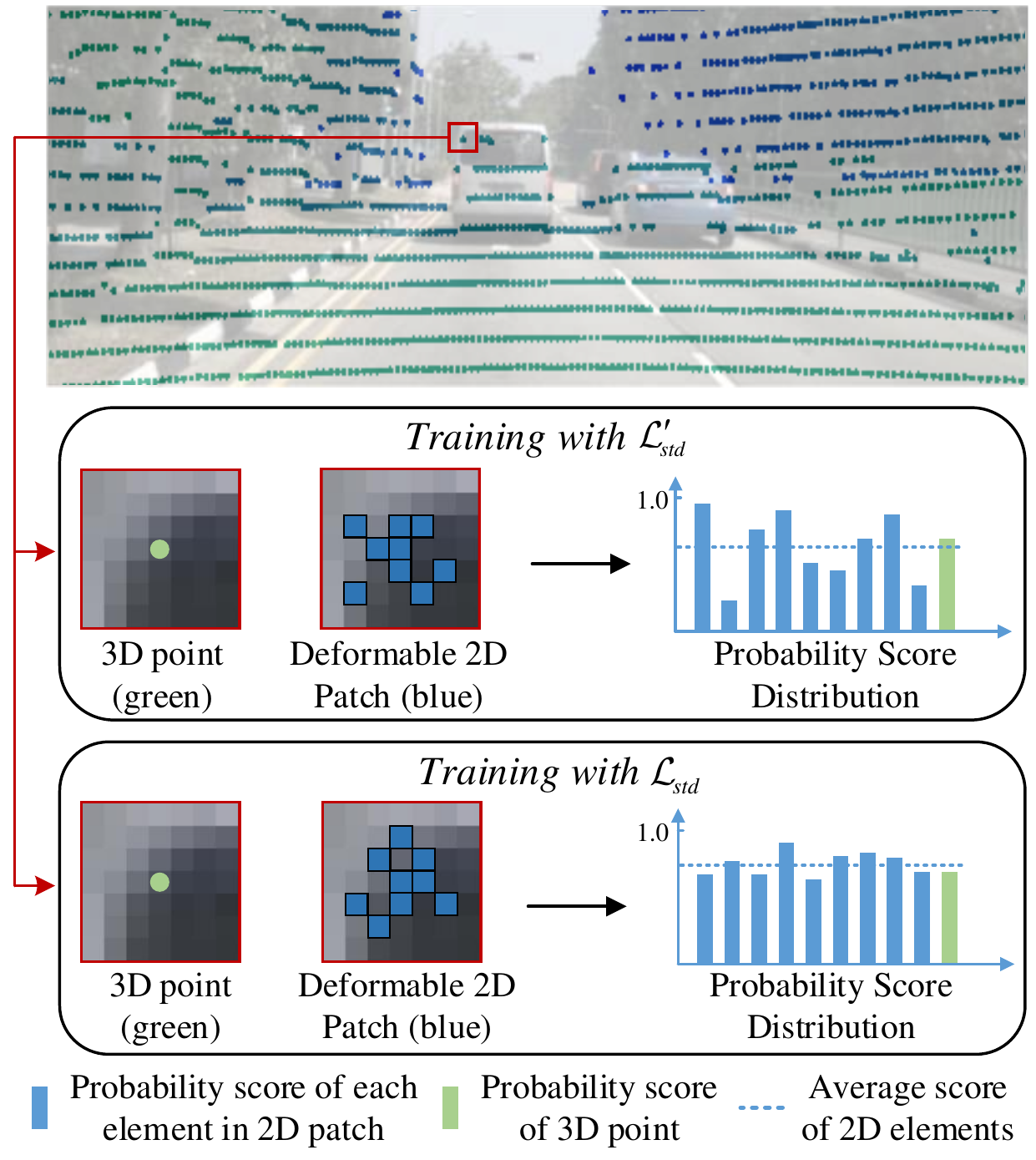} \caption{Comparision of probability distribution between learning with $\mathcal{L}_{std}^{\prime}$ and $\mathcal{L}_{std}$. We show actual probability scores of 2D patch and 3D point after training. As shown above, compared with our $\mathcal{L}_{std}$ which can effectively constrain each 2D element,  $\mathcal{L}_{std}^{\prime}$ may lead to elements with anomalous distribution despite favorable convergence of average distribution. }
    \label{--data-distribution}
    \vspace{-5mm}
    \end{figure}

    As mentioned in Sec. \ref{subsec:Intra-domain-cross-modal}, the sparse-to-dense loss $\mathcal{L}_{std}$ is proposed to constrain each element of a patch aiming to make them have a same probability distribution with the corresponding 3D point. It accomplishes this goal though constraining the supremum and infimum of each patch iteratively. Averaging is a straightforward way to integrate the information of all elements in patch, and naturally compress the 2D output to the same size of 3D output. One could be curious about whether the performance improvement could also be achieved if we change it to the loss based on mean value of each deformable patch. Specifically speaking, we address this concern by conducting additional experiments where the Eq. \ref{eq:std} is changed as follows:

    \begin{equation}
    \mathcal{L^{\prime}}_{std}=\frac{1}{N}\sum_{n=1}^{N}\mathcal{K}(\Phi_{avg}^{n}(\mathbf{P}_{2\mathrm{D}}),\mathbf{P}_{3\mathrm{D}}^{n}),
    \end{equation}
    where $\phi_{avg}^{n}(\mathbf{P}_{2D})$ denotes the average probability scores in the $n$-th 2D deformable patch. From the comparison results reported in Tab. \ref{tab:ablation-std}, we observe that with $\mathcal{L}_{std}^{\prime}$, DsCML performs at least 0.7\% worse than that with $\mathcal{L}_{std}$. The performance decline is obviously indicates the effectiveness of our Sparse-to-dense loss which constrains each element in patch.  As shown in Fig. \ref{--data-distribution}, constraining on average object of 2D patch is unable to ensure all elements are optimized to a same and correct direction. As mentioned before, $\mathcal{L}_{std}$ is the crucial loss function which enables cross modal learning in DsCML. To demonstrate the effectiveness of loss function between modalities. We conduct experiments using the model without $\mathcal{L}_{std}$. Results are shown the in last row of Tab. \ref{tab:ablation-std}. As the interaction between 2D and 3D is removed, it shows evident decline. This demonstrates that the loss between 2D and 3D is of great importance for domain adaptation. 

    \begin{table}
    \caption{Ablation on sparse-to-dense loss $\mathcal{L}_{std}$. $\mathcal{L^{\prime}}_{std}$ denotes the loss between 3D point and mean value of each deformable patch. Note that the decline ($\downarrow$) is obtained by comparing with the first model: DsCML($\mathcal{L}_{std}$).
    \label{tab:ablation-std}}
    \vspace{0.2cm}
    \centering{}\doublerulesep=0.5pt \resizebox{0.48\textwidth}{!}{%
    \begin{tabular}{l>{\centering}p{1.1cm}>{\centering}p{1.1cm}>{\centering}p{1.1cm}>{\centering}p{1.1cm}>{\centering}p{1.1cm}>{\centering}p{1.1cm}>{\centering}p{0.9cm}>{\centering}p{0.9cm}>{\centering}p{0.9cm}}
    \hline 
    \multirow{2}{*}{~~~~~~~Method} & \multicolumn{3}{c}{USA$\rightarrow$Singapore (nuScenes)} & \multicolumn{3}{c}{Day$\rightarrow$Night (nuScenes)} & \multicolumn{3}{c}{A2D2$\rightarrow$Sem.KITTI}\tabularnewline
    \cline{2-10}
     & 2D  & 3D & Avg & 2D  & 3D  & Avg & 2D  & 3D & Avg\tabularnewline
    \hline 
    \multirow{1}{*}{DsCML($\mathcal{L}_{std}$)} & 61.3  & 53.3  & 63.6 & 48.0  & 45.7  & 51.0 & 39.6  & 45.1 & 44.5\tabularnewline
    \hline 
    \multirow{2}{*}{DsCML($\mathcal{L^{\prime}}_{std}$)} & 60.2  & 52.5  & 62.9 & 47.1  & 44.9  & 50.3 & 37.6  & 44.3 & 43.6\tabularnewline
     & ($\downarrow$1.1)  & ($\downarrow$0.8)  & ($\downarrow$0.7) & ($\downarrow$0.9)  & ($\downarrow$0.8)  & ($\downarrow$0.7) & ($\downarrow$2)  & ($\downarrow$0.8) & ($\downarrow$0.9)\tabularnewline
    \hline 
    \multirow{2}{*}{DsCML(w/o $\mathcal{L}_{std}$)} & 53.4 & 46.9 & 61.5 & 41.8 & 41.6 & 47.6 & 36.5 & 37.3 & 42.3\tabularnewline
     & ($\downarrow$7.9)  & ($\downarrow$6.4)  & ($\downarrow$2.1) & ($\downarrow$6.2)  & ($\downarrow$4.1)  & ($\downarrow$3.4) & ($\downarrow$3.1)  & ($\downarrow$7.8) & ($\downarrow$2.2)\tabularnewline
    \hline 
    \end{tabular}}
    \vspace{-0.5cm}
    \end{table}

\section{Conclusion}

    In this paper, we present a multi-modality domain adaptation method for 3D semantic segmentation, which adopts both intra and inter domain cross modal learning. As for intra-domain CML, we advance Dynamic sparse-to-dense Cross Modal Learning (DsCML) to address the problem of insufficient information interaction between dense 2D and sparse 3D features. With the help of sparse-to-dense learning loss, DsCML builds effective consistency constraint between the two heterogeneous data. The design of deformable patch in DsCML enables the network to adaptively search the most suitable 2D region for knowledge transfer with 3D point. As for inter-domain CML, we utilize Cross Modal Adversarial Learning (CMAL) between output features which are both domain-different and modal-different aiming to introduce a higher level modality complementarity. Extensive experiments indicate that our approach achieves the superior performance to both uni-modal and multi-modal methods.

{\small
\bibliographystyle{ieee_fullname}
\bibliography{egbib}
}

\end{document}